\documentclass[11pt,a4paper]{article}
\usepackage[hyperref]{acl2020}
\usepackage{times}
\usepackage{latexsym}

\usepackage[utf8]{inputenc}
\usepackage{tikz}
 
\usepackage{amssymb}
 \usepackage{comment}
 \usepackage{booktabs}
 \usepackage{amsmath}
 \usepackage{arydshln}
 \usepackage{amsfonts}
 \usepackage{subfig}
 \usepackage{float}
\usepackage{url}

\usepackage{microtype}

\aclfinalcopy 

\title{Bootstrapping Disjoint Datasets for Multilingual Multimodal Representation Learning}

\author{Ákos Kádár \\
	Tilburg University \\
   {\tt a.kadar@uvt.nl}\\\And
      Grzegorz Chrupała\\
      Tilburg University \\
      {\tt g.chrupala@uvt.nl } \\\AND
      Afra Alishahi\\
      Tilburg University \\
      {\tt a.alishahi@uvt.nl}\\\And 
      Desmond Elliott \\
	University of Copenhagen \\
   {\tt de@di.ku.dk}\\
}

\date{}

\begin{document}
\maketitle

\begin{abstract}

Recent work has highlighted the advantage of jointly learning grounded sentence representations 
from multiple languages. 
However, the data used in these studies has 
been limited to an \emph{aligned} scenario:
the same images annotated with sentences 
in multiple languages. We focus on the more realistic 
\emph{disjoint} scenario in which there is no overlap between the images in multilingual image--caption datasets.
We confirm that training with aligned data results in better grounded sentence representations than training with \emph{disjoint} data, as measured by image--sentence retrieval performance.
In order to close this gap in performance, we propose a
\emph{pseudopairing} method to generate \emph{synthetically aligned} 
English--German--image triplets from the disjoint sets. The method works by
first training a model on the 
disjoint data, and then creating new triples across datasets using sentence similarity 
under the learned model.
Experiments show that pseudopairs improve
image--sentence retrieval performance compared to disjoint training, 
despite requiring no external data or models. 
However, we do find that using an external machine translation model 
to generate the synthetic data sets results in better performance.
\end{abstract}

\section{Introduction}

The perceptual-motor system plays an important role 
in concept acquisition and representation, 
and in learning the meaning of linguistic expressions
\citep{pulvermuller2005brain}. In natural language 
processing, many approaches have been proposed that
integrate visual information in the learning of 
word and sentence representations, highlighting
the benefits of visually grounded representations
\cite{lazaridou2015combining,baroni2016grounding,kiela2017learning,elliott2017imagination}.
In these approaches the visual world is
taken as a naturally occurring meaning representation for 
linguistic utterances, grounding language in perceptual
reality.

 Recent work has shown that we can learn better 
 visually grounded representations of sentences by
 training image--sentence ranking models on 
 multiple languages \cite{D17-1303,kadar2018conll}. 
 This line of research has
 focused on training models on datasets where the same 
 images are annotated with sentences in multiple languages. 
 This \emph{alignment} has
 either been in the form of the translation pairs 
 (e.g. German, English, French, and Czech in
 Multi30K \cite{W16-3210}) or independently collected
 sentences (English and Japanese in STAIR
 \cite{Yoshikawa2017}). 

  In this paper, we consider the problem of training an image--sentence ranking model using image-caption collections 
 in different languages with non-overlapping images drawn from different sources. We call these collections \emph{disjoint} datasets and argue that it is easier to find disjoint datasets than aligned datasets. This is especially the case for datasets in different languages, e.g. digital museum collections\footnote{Europeana: \url{http://www.europeana.eu/}}, newspaper collections \citep{ramisa2017breakingnews}, or the the images used in Wikipedia articles \citep{SchamoniHR18}. Multilingual aligned datasets, by contrast, are small and expensive to collect \citep{W16-3210}: there is a need for methods that can train image--sentence ranking models on disjoint multilingual image datasets.
 
 \newcite{kadar2018conll} claim that a multilingual image--sentence ranking model trained on disjoint datasets performs on-par with a model
 trained on aligned data. However, the disjoint datasets in their paper are artificial because they were formed by randomly splitting the Multi30K dataset into two halves. We examine whether the ranking model
 can benefit from multilingual supervision when it is
 trained using disjoint datasets drawn from different
 sources. In experiments with the Multi30K and COCO datasets,
 we find substantial benefits from training with these disjoint sources, but the best performance 
 comes from training on aligned datasets.
 
 Given the empirical benefits of training on aligned datasets, 
 we explore two approaches to creating synthetically 
 aligned training data in the \emph{disjoint} scenario. 
 One approach to creating synthetically aligned data is to use an
 off-the-shelf machine translation system to generate 
 new image-caption pairs by translating the original
 captions.
 This approach is very simple, but has the limitation 
 that an external system needs to be trained, which requires additional data. 
 
 The second approach is to generate synthetically aligned data that are \emph{pseudopairs}. We assume the existence of image--caption datasets in different languages where the images do not overlap between the datasets. Pseudopairs are created by annotating the images of one dataset with the captions from another dataset. This can be achieved by leveraging the sentence similarities predicted by an image-sentence ranking model trained on the original image--caption datasets. 
 One advantage of this approach is that it does not
 require additional models or datasets because it uses the 
 trained model to create new pairs. 
 The resulting pseudopairs can then be used 
 to re-train or fine-tune the original model. 

In experiments on the Multi30K and COCO datasets, we find that using an external machine translation system to create the synthetic data improves image--sentence ranking performance by 26.1\% compared to training on only the disjoint data. The proposed pseudopair approach consistently improves performance compared to the disjoint baseline by 6.4\%, and, crucially, this improvement is achieved without using any external datasets or pre-trained models. We expect that there is a broad scope for more complex pseudopairing methods in future work in this direction.
\section{Method}

We adopt the model architecture and training procedure
of \citet{kadar2018conll} for the task of matching images with sentences. This task is defined as learning to rank the sentences associated with an image higher than other sentences in the data set, and vice-versa \cite{hodosh2013framing}. The model is comprised of 
a recurrent neural network language model and a 
convolutional neural network image encoder. 
The parameters of the 
language encoder are randomly initialized, while
the image encoder is pre-trained, frozen during 
training and followed by a linear layer which is
tuned for the task.  
The model is trained to make true pairs $<a,b>$ 
similar to each other, and contrastive pairs
$<\hat{a},b>$ and 
$<a,\hat{b}>$ dissimilar from each other in a joint
embedding space by minimizing the max-violation loss function
\cite{faghri2017vse++}:
\begin{equation}
\label{eq:maxviol}
\begin{split}
\mathcal{J}(a, b) = &\max_{<\hat{a}, b>}[\text{max}(0, \alpha - \text{s}(a,b) + \text{s}(\hat{a}, b))] \;+ \\ &\max_{<a, \hat{b}>}[\text{max}(0, \alpha - \text{s}(a,b) + \text{s}(a, \hat{b}))]
\end{split}
\end{equation}

In our experiments, the $<a, b>$ pairs are either image-caption pairs 
$<i, c>$ or caption--caption pairs $<c_a, c_b>$ (following \newcite{D17-1303,kadar2018conll}).
When we train on $<i, c>$ pairs, we sample a batch
from an image--caption data set with uniform probability, encode the images and the sentences, and perform an update of the model parameters.
For the caption--caption objective, 
we follow \citet{kadar2018conll} and 
generate a sentence pair data set
by taking all pairs of sentences 
that belong to the same image
and are written in different languages: 5 English
and 5 German captions result in 25 English-German 
pairs. 
The sentences are encoded and we perform an update of 
the model parameters using the same loss. 
When training with both the image--caption and 
caption--caption (c2c) ranking tasks, 
we randomly select the task to perform with 
probability $p$=0.5. 

\subsection{Generating Synthetic Pairs}\label{sec:method:synthetic}

We propose two approaches to creating synthetic image--caption pairs to improve image--sentence ranking models when training with disjoint data sets. We assume the existence of datasets 
$\mathcal{D}_1$: $<\mathcal{I}^1$, $\mathcal{C}^{\ell_1}>$ and
$\mathcal{D}_2$: $<\mathcal{I}^2$, $\mathcal{C}^{\ell_2}>$
consisting of image--caption pairs $<i^1_i, c^{\ell_1}_i>$ and 
$<i^2_i, c^{\ell_2}_i>$ in 
languages $\ell_1$ and $\ell_2$, where the image sets do not overlap 
$\mathcal{I}^1 \cap \mathcal{I}^2 = \varnothing$. 
We seek to extend $<\mathcal{I}^2$, $\mathcal{C}^{\ell_2}>$ 
to a bilingual dataset with synthetic captions
$\hat{c}^{\ell_1}_i \in \hat{\mathcal{C}}^{\ell_1}$ 
in language $\ell_1$, resulting in a triplet data set 
$<\mathcal{I}^2$, $\hat{\mathcal{C}}^{\ell_1}$, $\mathcal{C}^{\ell_2}>$ 
consisting of triplets
 $<i^2_i, \hat{c}^{\ell_1}_i, c^{\ell_2}_i>$.
 We hypothesize that the new dataset will improve model performance because it will be trained to map the images to captions in both languages.

\subsection{Pseudopairs approach}\label{sec:method:pseudo}

Given two image-caption corpora  $<\mathcal{I}^1$, $\mathcal{C}^{\ell_1}>$  and  $<\mathcal{I}^2$, $\mathcal{C}^{\ell_2}>$ with pairs 
$<i^1_i, c^{\ell_1}_i>$ and $<i^2_i, c^{\ell_2}_i>$, we generate a \emph{pseudopair} corpus labeling each image in $\mathcal{I}^2$ with 
a caption from $\mathcal{C}^{\ell_1}$.
We create pseudopairs
only in one direction  leading 
to new image--caption pairs $<i^2, \hat{c}^{\ell_1}>$. 

The pseudopairs are generated using the sentence representations of
the model trained on both corpora  $<\mathcal{I}^1$, $\mathcal{C}^{\ell_1}>$
and $<\mathcal{I}^2$, $\mathcal{C}^{\ell_2}>$ jointly. 
We encode all 
captions $c^{\ell_1}_i \in \mathcal{C}^{\ell_1}$ and 
 $c^{\ell_2}_i \in \mathcal{C}^{\ell_2}$ and for each $c^{\ell_2}_i$
find the most similar caption $\hat{c}^{\ell_1}_i$ using the cosine similarity between the sentence representations.
This leads to pairs $<c^{\ell_2}_i, \hat{c}^{\ell_1}_i>$ 
and as a result to triplets $<i^2_i, c^{\ell_2}_i, \hat{c}^{\ell_1}_i>$ .

\paragraph{Filtering}

Optionally we filter 
the resulting pseudopair set $\mathcal{C}^{\ell_1}$, in an attempt to
avoid misleading samples with three filtering strategies:

\begin{enumerate}
    \item No filtering.
    \item Keep top: keep items with similarity scores in the 75\% percentile; keep top 25\% 
    \item Remove bottom: keep items with similarity scores in the 25\%; remove bottom 25\%
\end{enumerate}

\paragraph{Fine-tuning vs. restart}
After the pseudopairs are generated we consider two 
options: re-train the model from scratch with all previous 
data sets adding the generated pseudopairs or fine-tunening with
same data sets and the additional pseudopairs.

\subsection{Translation approach}\label{sec:method:mt}
Given a corpus  
$<\mathcal{I}^2$, $\mathcal{C}^{\ell_2}>$ with pairs 
$<i^2_i, c^{\ell_2}_i>$, we use a machine 
translation system 
to translate each caption $c^{\ell_2}_i$ to a language $\ell_1$ leading to new image--caption pairs 
$<i^2_i, \hat{c}^{\ell_1}_i>$\footnote{\citet{li2016adding} used a similar approach to create Chinese captions for images in the Flickr8K dataset, but they used the translations to train a Chinese image captioning model.}. 
Any off-the-shelf translation system could be used to create the translated captions, e.g. an online service or a pre-trained translation model.
\section{Experimental Protocol}

\subsection{Model}

Our implementation, training protocol and parameter settings are based on the existing codebase of \newcite{kadar2018conll}.\footnote{\url{https://github.com/kadarakos/mulisera}}.
In all experiments, we use the 2048 dimensional image features extracted from the
last average-pooling layer of a pre-trained\footnote{Trained on the ILSVRC 2012 1.2M image 1000 class object classification
subset ImageNet \cite{russakovsky2015imagenet}} 
ResNet50 CNN \cite{he2016deep}. 
  
The image representation used in our model is obtained by a single 
affine transformation that we train from scratch 
$\mathbf{W}_{I} \in \mathbb{R}^{2048 \times 1024}$. For the sentence
encoder we use a uni-directional Gated Recurrent Unit (GRU) 
network \cite{cho2014properties} with a single hidden layer with 
1024  hidden units and 300 dimensional word embeddings. 
When training bilingual models we use a single word embedding
for the same word-forms, making no distinction if they come 
from different languages. Each sentence is represented by the final 
hidden state of the GRU. For the similarity function in the loss function (Eq. 1) we use cosine similarity and $\alpha=0.2$ margin parameter.

In all experiments, we early stop on the validation set when no improvement is observed for 10 inspections, which are performed every 500 updates. 
The stopping criterion is the sum of text-to-image (T$\rightarrow$I) and 
image-to-text (I$\rightarrow$T) recall scores at ranks 
1, 5 and 10 across all languages in the training data. 
The models are trained with a batch-size of 128 with with the Adam optimizer \cite{kingma2014adam} 
using default parameters and an initial learning rate of
\mbox{2e-4}
without applying any learning-rate decay schedule.
We apply gradient norm clipping with a value of 2.0.

We use a pre-trained OpenNMT \cite{2017opennmt} English-German machine translation model\footnote{\footnotesize{\url{https://s3.amazonaws.com/opennmt-models/wmt-ende_l2-h1024-bpe32k_release.tar.gz}}} to create the data for the translation approach described in Section \ref{sec:method:mt}.

\subsection{Datasets}

The models are trained and evaluated on the bilingual English-German Multi30K dataset (M30K), and we
optionally train on the English COCO dataset 
\cite{DBLP:journals/corr/ChenFLVGDZ15}. In \textit{monolingual} 
experiments, the model is trained on a single language from M30K or 
COCO.

In the \emph{aligned} bilingual experiments, we use the
independently collected English and German captions in M30K:
The training set consists of 29K images and 145K captions; the validation and test sets have 1K images and 5K captions. 

For the \emph{disjoint} experiments, we use the COCO data set
with the \newcite{karpathy2015deep} splits. This gives 82,783 training,
5,000 validation, and 5,000 test images; each image is paired with five captions.
The data set has an additional split containing the  
30,504 images from the original validation set of 
MS-COCO (``restval''),
 which we add to the training set as in previous work \cite{karpathy2015deep,vendrov2016order,faghri2017vse++}.
 

\subsection{Evaluation}
We report results on Multimodal Translation Shared Task 2016 test split \cite{specia2016shared} of M30K. 
Due to space constraints, we only report recall at 1 (R@1) for 
Image-to-Text (I$\rightarrow$T) and Text-to-Image (T$\rightarrow$I)
retrieval, and the sum of R@1, R@5, and R@10 recall scores across both tasks and languages (Sum).\footnote{This is the criterion we use
for early-stopping.}
\section{Baseline Results}\label{sec:results}

The experiments presented here set the baseline performance 
for the visually grounded bilingual models
and introduces the data settings that we will use in the 
later sections.

\paragraph{Aligned}

\begin{table}[t]
    \centering
    \renewcommand{\arraystretch}{1.0}
    \begin{tabular}{lccc}
        \toprule
         & \multicolumn{3}{c}{English}\\
         \cmidrule(lr){2-4}
         & I $\rightarrow$ T & T $\rightarrow$ I & Sum\\
         \midrule
         En & 40.5 & 28.8 & 346.4 \\
         \; + De & 41.4 & 29.9  & 352.8\\
         \; \; + c2c & 42.8  & 32.1 & 361.6\\
         \hdashline
         COCO & 34.4 & 24.8 & 304.0 \\
         \; + En & \textbf{46.2} & \textbf{33.4} & \textbf{374.4}\\
         \bottomrule
    \end{tabular}
    \caption{Performance on the English
    M30K 2016 test set in the \emph{aligned setting} for models trained on 
    M30K English (En), both M30K German and English 
    (+De), with caption ranking (+c2c), 
    COCO (COCO) and both COCO and M30K English (+En).}
    \label{tab:engbaseline}
\end{table}

\begin{table}
    \centering
    \renewcommand{\arraystretch}{1.0}   
    \begin{tabular}{lccc}
        \toprule
         & \multicolumn{3}{c}{German} \\
         \cmidrule(lr){2-4}
         & I $\rightarrow$ T & T $\rightarrow$ I & Sum\\
         \midrule
         De & 34.9 & 24.6 & 311.2 \\
         \; + En & \textbf{38.6} & 26.0 & 324.6\\
         \; \; + c2c & 38.3 & \textbf{27.7} &   \textbf{334.0} \\
         \; + COCO & 36.4 & 25.7 & 319.7\\
         \bottomrule
    \end{tabular}
        \caption{Performance on the German
    M30K 2016 test set in the \emph{aligned} and
    \emph{disjoint} settings for models trained on 
    M30K German (De), both M30K German and English (+En), 
    and with caption ranking (+c2c) and both M30K German and COCO (+COCO).}
    \label{tab:gerbaseline}
\end{table}

\begin{table*}[h!]
    \centering
    \renewcommand{\arraystretch}{1.0}
    \begin{tabular}{lcccccc}
        \toprule
         & \multicolumn{3}{c}{English} & \multicolumn{3}{c}{German}\\
        \cmidrule(lr){2-4} \cmidrule(lr){5-7}
         & I $\rightarrow$ T & T $\rightarrow$ I & Sum & I $\rightarrow$ T & T $\rightarrow$ I & Sum \\
         \midrule
         En+De+c2c & 42.8 & 28.6 & 361.6 & 38.3 & 27.7 & 334.0 \\
         \; + COCO & \textbf{46.5} & \textbf{34.8} &  \textbf{378.9}  & \textbf{40.6} & \textbf{28.8} & \textbf{344.6}\\
         \bottomrule
    \end{tabular}
    \caption{Recall @ 1 and Sum-of-Recall-Scores for Image-to-Text (I $\rightarrow$ T) and Text-to-Image (T $\rightarrow$ I) baseline results on the English and German
    M30K 2016 test in the \emph{aligned plus disjoint} setting}\label{tab:alignedplus}
\end{table*}

In these experiments we only use the \emph{aligned} English-German data from M30K. 
Tables~\ref{tab:engbaseline} and \ref{tab:gerbaseline} present the result for English and German, respectively. The Sum-of-recall scores for both languages show that the best approach is the bilingual model with the c2c loss (En+De+c2c, and De+En+c2c). These results reproduce the findings of
\newcite{kadar2018conll}.

\paragraph{Disjoint}\label{sec:baseline_results:disjoint}

We now determine the performance of the model when it is trained on data drawn from different data sets with no overlapping images.

First we train two English \emph{monolingual}
models: one on the M30K English dataset and one on the English COCO dataset. Both models are evaluated on image--sentence ranking performance on the M30K English test 2016 set. The results in Table~\ref{tab:engbaseline} show that there is a substantial difference in performance in both text-to-image and image-to-text retrieval, depending on whether the model is trained on the M30K or the COCO dataset.
The final row of Table~\ref{tab:engbaseline} shows, however, 
that jointly training on both data sets improves over 
only using the M30K English training data.

We also conduct experiments in the \emph{bilingual disjoint} setting, 
where we study whether it is possible to improve the performance of a German model using the
out-of-domain English COCO data. Table~\ref{tab:gerbaseline}
shows that there is an increase in performance when the model is trained on the disjoint sets, as opposed to
only the in-domain M30K German (compare De against De+COCO). This result is not too surprising as we have observed both the advantage of joint training on both languages in the \emph{aligned} setting and the overlap between the different datasets. 

Finally, we compare the performance of a German model trained in the \emph{aligned}
and \emph{disjoint} settings. We find that a model trained in the \emph{aligned}
setting (De+En) is better than a model trained in the \emph{disjoint} setting (De+COCO), as shown in Table \ref{tab:gerbaseline}.
This finding contradicts the conclusion of \citet{kadar2018conll}, who claimed that the
\emph{aligned} and \emph{disjoint} conditions lead to comparable performance. This is most likely because the disjoint setting in \newcite{kadar2018conll} is artificial, in the sense that they used different 50\% subsets of M30K. In our experiments the disjoint image--caption sets are real, in the sense that we trained the models on the two different datasets.
 
\paragraph{Aligned plus disjoint}

Our final baseline experiments explore the combination of 
\emph{disjoint} and \emph{aligned} data settings. We train an English-German bilingual model with the c2c
objective on M30K, and we also train on the English COCO data. Table~\ref{tab:alignedplus} shows that adding the 
\emph{disjoint} data improves performance for
both English and German compared to training 
solely the \emph{aligned} model. 

\paragraph{Summary}
First we reproduced the findings of \citet{kadar2018conll} 
showing that bilingual joint
training improves over monolingual and using c2c loss
further improves performance. 
Furthermore, we have found that adding the COCO as 
additional training data both when only training on German, and training on both German-English from 
M30K improves performance 
even if the model is trained on data drawn from a different dataset.

\begin{table}[t]
    \centering
    \renewcommand{\arraystretch}{1.0}
    \begin{tabular}{lccc}
        \toprule
         & \multicolumn{3}{c}{German}\\
         & I $\rightarrow$ T & T $\rightarrow$ I & Sum\\
         \midrule
         De + COCO &  36.4  & 25.7 & 319.7\\
         \: \: + pseudo &  37.3 & 25.2 &  319.9\\
          \: \: \: + fine-tune &  \textbf{38.0} & 25.6 & \textbf{322.9}\\
          \: \: + pseudo 25\% &  37.3 & \textbf{25.9}  &  320.9\\
          \: \: \: + fine-tune &  37.2 & 25.7 &  320.7 \\
          \: \: + pseudo 75\% & 36.8  & 25.1 &  316.3\\
          \: \: \: + fine-tune &  36.5  & 25.5 &  317.5 \\

         \bottomrule
    \end{tabular}
    \caption{A \emph{disjoint} model is trained on the De+COCO datasets and used
    to generate pseudopairs. Then the full pseudopair set (+pseudo) or the filtered versions (+pseudo 25\% and +pseudo 75\%) are used as an extra data set to either re-train the moddel from scratch or fine-tune the original De+COCO model (+fine-tune).}
    \label{tab:gerpseudodisjoint}
\end{table}

\begin{table*}[ht]
    \centering
    \renewcommand{\arraystretch}{1.0}
    \begin{tabular}{lccccccc}
        \toprule
         & \multicolumn{3}{c}{English}  & \multicolumn{3}{c}{German}\\
         \cmidrule(lr){2-4} \cmidrule(lr){5-7}
         & I $\rightarrow$ T & T $\rightarrow$ I & Sum 
         & I $\rightarrow$ T & T $\rightarrow$ I & Sum & Sum(Sum)\\
         \midrule
         En+De+COCO+c2c & 46.5 & 34.8 & 378.9 & 40.6 & 28.8 & 344.6 & 723.5\\
         \: \: + pseudo & \textbf{48.1}  & 35.6 & \textbf{382.3} & \textbf{41.8}  & 29.0  & 345.6 & 727.8 \\
          \: \: \: + fine-tune & 47.0 & \textbf{35.7}  & 381.5  & 40.9  & 28.7 & 346.8 & \textbf{728.2} \\
          \: \: + pseudo 25\% &  47.5  & 34.9 &  380.2 &  41.5 & 28.9 &  345.5 & 725.7\\
           \: \: \: + fine-tune &  46.1 & 35.4 &  379.7  &  41.6 & \textbf{29.1} & \textbf{347.8} & 727.5\\
          \: \: + pseudo 75\% &  45.9  & 34.0 & 373.6  & 40.3  & 27.9 &  339.1 & 712.7\\
          \: \: \: + fine-tune & 46.2  & 35.1 & 378.6  &  41.0 & \textbf{29.1}  &  345.1 & 723.6\\
          \hdashline
          \: + Translation & \textbf{47.5} & \textbf{36.2} & \textbf{384.5} & \textbf{43.5} & \textbf{30.5} & \textbf{357.9} & \textbf{742.4}\\
         \bottomrule
    \end{tabular}
    \caption{We train the \emph{aligned plus disjoint} model with 
    c2c loss and add the full pseudopair set (+pseudo) or the filtered versions (+pseudo 25\% and +pseudo 75\%) is added as an extra data set. The model is either re-trained from scratch or fine-tuned (+fine-tune). We also report the result of training the \emph{aligned plus disjoint} model with the synthetic translations (+Translation).}
    \label{tab:pseudoalignedplus}
\end{table*}

\section{Training with Pseudopairs}
\label{sec:pseudo}

In this section we turn our attention to creating a synthetic English-German \emph{aligned} data set from the English COCO using the pseudopair method (Section \ref{sec:method:synthetic}). The synthetic data set is used to train an image-sentence ranking model either from scratch or by fine-tuning the original model; in addition, we also explore the effect of using all of the pseudopairs or by filtering the pseudopairs. We hypothesise that training a model with the additional pseudopairs with improve over the \textit{aligned plus disjoint} baseline.

\paragraph{Disjoint}

We generate pseudopairs using the 
\emph{disjoint} bilingual model trained on 
the German M30K and the English COCO.
Table~\ref{tab:gerpseudodisjoint} reports the results
when evaluating on the M30K German data.
Line 2 shows that using the full pseudopair set
and re-training the model does not lead to 
noticeable improvements. However, line 3 shows that performance increases when we train with all pseudopairs and fine-tuning the original disjoint bilingual model. Filtering the pseudopairs at either the 25\% and 75\% percentile is 
detrimental to the final performance.\footnote{
We did not find any improvements
in the \emph{disjoint} setting when training with pseudopairs and the additional c2c loss.}

\paragraph{Aligned plus disjoint}

We generate pseudopairs using a model trained on M30K English-German data with the c2c objective and the English
COCO data set. The results for both English and German are reported in Table~\ref{tab:pseudoalignedplus}; note that when we train with the pseudopairs we also train with the c2c loss on both data sets.  
Overall we find that pseudopairs 
improve performance, however, we do not achieve the 
best results for English and German in the same conditions. The best results for German are to filter at 25\% percentile and apply fine-tuning, 
while for English the best results are without filtering or fine-tuning. The best overall model is trained with
all the pseudopairs with fine-tuning, according to the Sum of the Sum-of-recall scores across 
both English and German.
The performance across both data sets is increased from 723.5 to 728.2 using the pseudopair method.

\paragraph{Summary}
In both \emph{aligned plus disjoint} and \emph{disjoint} scenarios, the 
additional pseudopairs improve performance, and in both cases the overall 
best performance is achieved when applying the 
fine-tuning strategy and no filtering of the samples.
\begin{table}[t]
    \centering
    \renewcommand{\arraystretch}{1.0}
    \begin{tabular}{lccc}
        \toprule
         & \multicolumn{3}{c}{German}\\
         & I $\rightarrow$ T & T $\rightarrow$ I & Sum\\
         \midrule
         De + COCO & 36.4 & 25.7 & 319.7\\
         \: \: + Translation & 37.7 & 26.3 & 327.2\\
         \: \: \: + c2c      & {\bf 39.9} & {\bf 26.7} & {\bf 335.5}\\
         \bottomrule
    \end{tabular}
    \caption{Results on the German
    M30K 2016 test set with the \emph{aligned plus disjoint} 
    (En+De+COCO+c2c) model, the additional automatically translated COCO (+Translation) and with the c2c on the synthetic pairs.
    }
    \label{tab:german_translation_disjoint}
\end{table}

\section{Training with Translations}\label{sec:transaltion}

We now focus on our second approach to creating an 
English-German \textit{aligned} dataset 
using the translation method described in Section \ref{sec:method:synthetic}.

\paragraph{Disjoint}

We first report the results of disjoint bilingual
model trained on the German M30K, the English COCO data, 
and the translated German COCO in Table \ref{tab:german_translation_disjoint}. 
The results show that retrieval performance is improved when the model is trained on the translated German 
COCO data in addition to the English COCO data. 
We find the best performance when we jointly train on the 
M30K German, the Translated German COCO and 
the English COCO with the additional c2c objective 
over the COCO datasets (+c2c). 
We note that this setup leads to a better model, as measured by the sum-of-recall-scores, than training on the aligned M30K data (compare De+COCO+Translation+c2c in Table \ref{tab:german_translation_disjoint} to De+En+c2c in Table \ref{tab:gerbaseline}).

\paragraph{Aligned plus Disjoint}

In these experiments, we train models with the {\it aligned} M30K  
data, the {\it disjoint} English COCO data, and the translated German COCO data. 
Table \ref{tab:pseudoalignedplus} 
presents the results for the English and German evaluation. 
We find that training on the German Translated COCO data and 
using the c2c loss over the COCO data results in improvements for both languages. 

\paragraph{Summary}

In both the \emph{disjoint} and 
\emph{aligned plus disjoint} settings, we find that training 
with the translations of COCO improves 
performance over training with only the English COCO data. 

\section{Discussion}

\subsection{Sentence-similarity quality}

The core of the proposed pseudopairing method is based on measuring the similarity between sentences, but how well does our model encode similar sentences? Here we analyze the ability of our models to identify translation equivalent sentences using the English-German translation pairs in the Multi30K test 2016 data. This experiment proceeds as follows: (i) we assume a pre-trained image--sentence ranking model, (ii) we encode the German and English sentences using the language encoder of the model, (iii) we calculate the model's performance on the task of ranking the correct translation for English sentences, given the German caption, and vice-versa. 

To put our results into perspective we compare to the 
best approach
to our knowledge as reported by \newcite{rotman2018bridging}: DPCCA is a deep partial canonical correlation analysis method maximizing the canonical correlation between captions of the same image conditioned on image representations as a third view. 
Table~\ref{tab:translate} reports the results of this experiment.
Our models consistently improve upon the state-of-the-art. The baseline aligned model trained on the Multi30K data slightly outperforms the DPCCA for EN $\rightarrow$ DE retrieval, and more substantially outperforms DPCCA for DE $\rightarrow$ EN. If we train the same model with the additional c2c objective, R@1 improves by 8.0 and 12.1 points, respectively. We find that adding more monolingual English data from the external COCO data set slightly degrades retrieval performance, and that performing sentence retrieval using a model trained on the disjoint M30K German and English COCO data sets result in much lower retrieval performance. We conclude that the model that we used to estimate sentence similarity is the best-performing method known for this task on this data set, but there is room for improvement for models trained on disjoint data sets.

\begin{table}[]
    \centering
    \begin{tabular}{lcc}
    \cmidrule[0.08em](r){1-3}
    & EN {\small $\rightarrow$} DE & DE {\small $\rightarrow$} EN \\
    \cmidrule(rl){1-3}
    DPCCA &  82.6 & 79.1 \\
    \cmidrule(rl){1-3}
    En + De & 82.7   & 83.4  \\
    En + De + c2c & \textbf{90.6} & \textbf{91.2}  \\
    En + De + COCO & 82.5    & 81.0    \\
    En + De + COCO + c2c & 90.0   & 90.1  \\
    De + COCO & 73.4 &  70.7  \\
    \cmidrule[0.08em](r){1-3}
    \end{tabular}
    \caption{Translation retrieval results (Recall @ 1) on the M30K 2016 test set compared to the state of the art.}
    \label{tab:translate}
\end{table}

\subsection{Characteristics of the Pseudopairs}

We now investigate the properties of the pseudopairs generated by our method. In particular, we focus on pseudpairs generated by an aligned plus disjoint model (En+De+COCO+c2c) and a disjoint model (De+COCO).

The pseudopairs generated by the \textit{aligned plus disjoint} model cover 40\% of the German captions in the M30K data set, and overall, the pseudopairs form a heavy-tailed distribution.
We find a similar pattern for the pseudopairs generated by the \textit{disjoint} model: the pseudopairs cover 37\% of the M30K data set, and the top 150 captions cover 23\% of the data. This is far from using each caption equally in the pseudopair transfer, and may suggest a hubness problem \cite{dinu2014improving}. We assessed the stability of the sets of transferred captions using the Jaccard measure in two cases: (i) different random seeds, and (ii) disjoint or aligned plus disjoint. For the \emph{aligned plus disjoint} model, we observe an overlap of 0.53 between different random seeds compared to 0.51 for the \emph{disjoint} model. The overlap between the two types of models is much lower at 0.41. Finally, we find that when a caption is transferred by both models, the overlap of the caption annotating the same COCO image is 0.33 for the \emph{disjoint} model, and 0.34 for the \emph{aligned plus disjoint} model, and the overlap between the models is 0.16. This shows that the models do not transfer the same captions for the same images.


Figure \ref{fig:pseudopairs} presents examples of the annotations transferred using the pseudopair method. The first example demonstrates the difference between the Multi30K and COCO datasets: there are no giraffes in the former, but there are dogs (``Hund'').  In the second example, both captions imply that the man sits \emph{on} the tree not beside it. This shows that even if the datasets are similar, transferring a caption that exactly matches the picture is difficult. The final two examples show semantically accurate and similar sentences are transferred by both models. In the fourth example, both models transfer exactly the same caption.

\DeclareRobustCommand{\circled}[1]{%
	\tikz[baseline=(char.base)]{
            \node[shape=circle,draw,inner sep=2pt] (char) {#1};}}

\begin{figure*}[t]
\begin{center}

\setlength\tabcolsep{8pt}
\begin{tabular}{lcccc}
\subfloat[\circled{1} \vspace{5.2em}\newline \circled{2}] &
\subfloat[Ein hund steht auf einem baumstamm im wald. \newline$\lbrack$\emph{A dog is standing on a tree trunk in the forest.}$\rbrack$ \vspace{3.2em}\newline Hund im wald. \newline$\lbrack$\emph{Dog in the forest.}$\rbrack$]
{\includegraphics[width = 0.21\textwidth]{assets/8.png}} &
\subfloat[Mann sitzt im baum. \newline$\lbrack$\emph{Man is sitting in the tree.}$\rbrack$ \vspace{4.2em}\newline\newline Der mann der auf einem baum sitzt. \newline$\lbrack$\emph{The man sits on the tree.}$\rbrack$]
{\includegraphics[width = 0.21\textwidth]{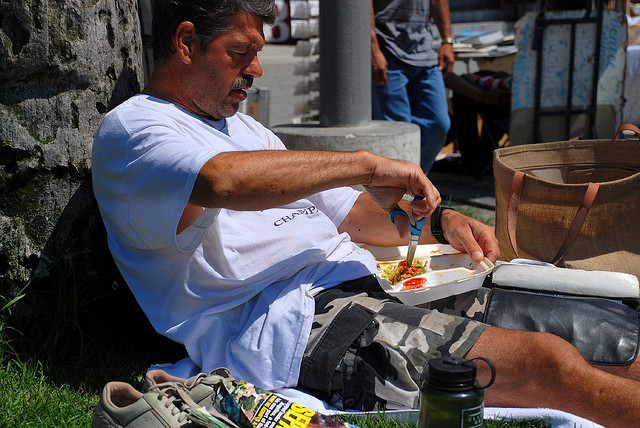}} &
\subfloat[Ein mann sitzt in einem boot auf einem see. \newline$\lbrack$\emph{A man is sitting in a boat on a lake.}$\rbrack$ \vspace{3.2em}\newline Ein mann sitzt am see auf dem ein boot fährt. \newline$\lbrack$\emph{A man is sitting at the lake on which a boat is riding.}$\rbrack$]
{\includegraphics[trim={0 0 0 2.3cm}, clip, width = 0.21\textwidth]{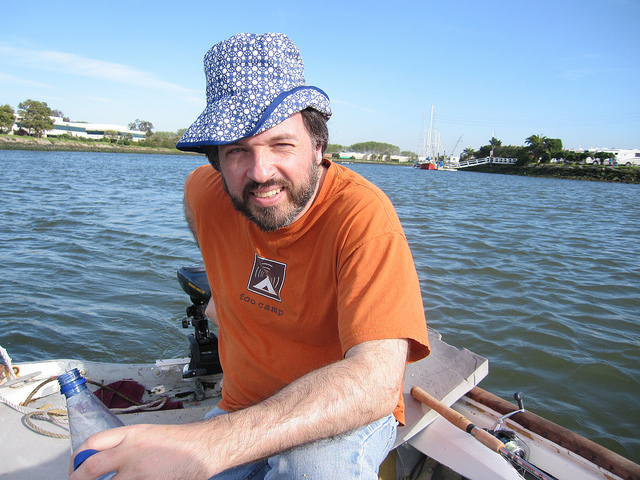}} &
\subfloat[Ein jet jagt steil in die luft, viel rauch kommt aus dem rumpf. \newline$\lbrack$\emph{A jet goes steep up into the air, a lot of smoke is coming out of its hull.}$\rbrack$ \vspace{1em}\newline Ein jet jagt steil in die luft, viel rauch kommt aus dem rumpf. \newline$\lbrack$\emph{A jet goes steep up into the air, a lot of smoke is coming out of its hull.}$\rbrack$]
{\includegraphics[width = 0.21\textwidth]{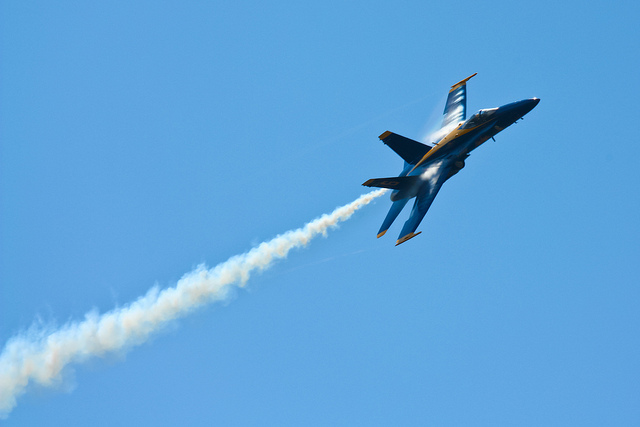}}
\end{tabular}
\end{center}

\caption{Visualisation of the sentences transferred from Multi30K to the COCO data set using the pseudopair method. \circled{1} is transferred from a model trained on De+COCO, whereas \circled{2} is transferred from En+De+COCO. $\lbrack$English glosses of the sentences are included for ease of reading.$\rbrack$}
\label{fig:pseudopairs}
\end{figure*}

\section{Related Work}


Image--sentence ranking is the task of retrieving the sentences that best describe an image, and vice-versa \cite{hodosh2013framing}. Most recent approaches are based on learning to project image representations and sentence representations into a shared space using deep neural networks \cite[\textit{inter-alia}]{frome2013devise,socher2014grounded,vendrov2016order,faghri2017vse++}. 

More recently, there has been a focus on solving this task using multilingual data \cite{D17-1303,kadar2018conll} in the Multi30K dataset \cite{W16-3210}; an extension of the popular Flickr30K dataset into German, French, and Czech.
These works take a multi-view learning perspective in which 
images and their descriptions in multiple languages are different views of the
same concepts. The assumption is that common representations of multiple languages and perceptual stimuli 
can potentially exploit complementary information
between views to learn better representations. 
For example, \newcite{rotman2018bridging} improves bilingual sentence representations by incorporating image information as a third view by Deep Partial Canonical Correlation Analysis.
More similar to our work 
\newcite{D17-1303}, propose a convolutional-recurrent architecture with both an image--caption
and caption--caption loss to learn bilingual visually grounded representations. 
Their results were improved by the approach presented in \newcite{kadar2018conll}, who has also
shown that the multilingual models outperform bilingual models, and that image--caption retrieval 
performance in languages with less resources can be improved with data from higher-resource
languages. We largely follow \citet{kadar2018conll}, however, our main interest lies in learning multimodal
and bilingual representations in the scenario where the images do not come from the same
data set i.e.: the data is presented is two sets of image--caption tuples rather than
image--caption--caption triples.

Taking a broader perspective, images have been used as pivots in multilingual multimodal language processing. On the word level this intuition is 
applied to visually grounded bilingual lexicon induction, which aims to learn 
cross-lingual word representations without aligned text using images as pivots
\cite{bergsma2011learning,kiela2015visual,vulic2016multi,hartmann2017limitations,hewitt2018learning}. Images have been used as pivots to learn translation models only from image--caption
data sets, without parallel text \cite{hitschler2016multimodal,nakayama2017zero,lee2017emergent,chen2018zero}.
\section{Conclusions}

Previous work has demonstrated improved image--sentence ranking performance
when training models jointly on multiple languages \cite{D17-1303, kadar2018conll}. Here we presented a study on learning multimodal
and multilingual representations in the \emph{disjoint setting}, where
images between languages do not overlap.
We found that learning visually grounded sentence 
embeddings in this
setting is more challenging. 
To close the gap, we developed a \emph{pseudopairing} 
technique that creates synthetic pairs by annotating the
images from one of the data sets with the image 
descriptions of the other using the
sentence similarities of the model trained on both. 
We showed that training with pseudopairs 
improves performance,
without the need to augment training from additional 
data sources or other pipeline components. 
However, our technique is outperformed
by creating synthetic pairs using an off-the-shelf automatic machine translation system. 
As such our results suggest that it is better to 
use translation, when a good translation system 
is available, however, in its absence, pseudopairs
offer consistent improvements. 
We have found that our pseudopairing method only transfers annotations from a small number of images and in the future we plan to substitute our naive matching algorithms
with approaches developed to mitigate this hubness issue \cite{radovanovic2010existence} and to close the gap between translation and pseudopairs.

\bibliography{bibliography}
\bibliographystyle{acl_natbib}

\end{document}